\documentclass{article} 
\usepackage{iclr2023_conference,times}

\usepackage{amsmath,amsfonts,bm}

\def\eqref#1{equation~\ref{#1}}

\def\1{\bm{1}}

\DeclareMathAlphabet{\mathsfit}{\encodingdefault}{\sfdefault}{m}{sl}
\SetMathAlphabet{\mathsfit}{bold}{\encodingdefault}{\sfdefault}{bx}{n}

\usepackage{hyperref}
\usepackage{url}
\usepackage{graphicx}
\usepackage{microtype}
\usepackage{subfigure}
\usepackage{booktabs} %
\usepackage{multirow}
\usepackage{makecell}
\usepackage{capt-of}
\usepackage{hyperref}
\hypersetup{   
    urlcolor=blue,
}
\usepackage{bbm}

\def\laion{LAION 400M dataset}

\usepackage{enumitem}
\newenvironment{tight_itemize}{
\begin{itemize}[leftmargin=20pt]
  \setlength{\topsep}{0pt}
  \setlength{\itemsep}{0pt}
  \setlength{\parskip}{0pt}
  \setlength{\parsep}{0pt}
}{\end{itemize}}

\title{Unicom: Universal and Compact \\ Representation Learning for Image Retrieval}

\author{\\{\bf Xiang An}$^{1}$,~{\bf Jiankang Deng}$^{2}$ \thanks{denotes corresponding author.}~~,~{\bf Kaicheng Yang}$^{1}$,~{\bf Jiawei Li}$^{1}$,~{\bf Ziyong Feng}$^{1}$,\\{\bf Jia Guo}$^{3}$,~{\bf Jing Yang}$^{4}$,~{\bf Tongliang Liu}$^{5}$  \\
$^{1}$DeepGlint, $^{2}$Huawei, $^{3}$InsightFace, $^{4}$University of Cambridge, $^{5}$University of Sydney\\
\texttt{xiangan@deepglint.com,jiankangdeng@gmail.com} \\}

\iclrfinalcopy 
\begin{document}

\maketitle

\begin{abstract}
Modern image retrieval methods typically rely on fine-tuning pre-trained encoders to extract image-level descriptors.
However, the most widely used models are pre-trained on ImageNet-1K with limited classes. The pre-trained feature representation is therefore not universal enough to generalize well to the diverse open-world classes. 
In this paper, we first cluster the large-scale \laion{} into one million pseudo classes based on the joint textual and visual features extracted by the CLIP model. Due to the confusion of label granularity, the automatically clustered dataset inevitably contains heavy inter-class conflict. To alleviate such conflict, we randomly select partial inter-class prototypes to construct the margin-based softmax loss. To further enhance the low-dimensional feature representation, we randomly select partial feature dimensions when calculating the similarities between embeddings and class-wise prototypes. The dual random partial selections are with respect to the class dimension and the feature dimension of the prototype matrix, making the classification conflict-robust and the feature embedding compact. Our method significantly outperforms state-of-the-art unsupervised and supervised image retrieval approaches on multiple benchmarks. The code and pre-trained models are released to facilitate future research \href{https://github.com/deepglint/unicom}{\textcolor{blue}{https://github.com/deepglint/unicom}}. 

\end{abstract}

\section{Introduction}
Modern image retrieval methods \citep{Lim2022CVPR,Roth2022CVPR,Kim2022CVPR,Ermolov_2022_CVPR,Patel2022CVPR} can be roughly decomposed into two major components: (1) the encoder (e.g., Convolutional
Neural Networks \citep{szegedy2015going,he2016deep} or Vision Transformer \citep{touvron2021training,dosovitskiy2021image}) mapping the image to its compact representation and (2) the loss function \citep{musgrave2020metric} grouping the representations of similar objects while pushing away representations of dissimilar objects in the embedding space.  
To train the encoder, networks pre-trained on crowd-labeled datasets (e.g., ImageNet \citep{deng2009imagenet}) are widely used for fine-tuning \citep{wang2019multi,kim2021embedding}. However, ImageNet only contains 1,000 pre-defined object classes. The feature representation learned from ImageNet is not universal enough to generalize to diverse open-world objects.

Even though fully supervised pre-training can benefit from a strong semantic learning signal for each training example, supervised learning is not scalable because manual annotation of large-scale training data is time-consuming, costly, and even infeasible. 
By contrast, self-supervised pre-training methods \citep{he2020momentum,he2022masked,radford2021learning,jia2021scaling} can be easily scaled to billions of unlabeled examples by designing an appropriate pretext task, such as solving jigsaw puzzles \citep{noroozi2016unsupervised}, invariant mapping \citep{chen2021exploring}, and image-text matching \citep{radford2021learning,jia2021scaling}. Among them, \textcolor{black}{CLIP \citep{radford2021learning} has recently demonstrated success across various downstream tasks (e.g., image retrieval and classification) due to superior feature representation empowered by image-text contrastive learning. Specifically, CLIP aligns the visual and textual signals of each instance into a unified semantic space by cross-modal instance discrimination.}
Nevertheless, the instance discrimination method used by CLIP can hardly encode the semantic structure of training data, because instance-wise contrastive learning always treats two samples as a negative pair if they are from different instances, regardless of their semantic similarity. 
When thousands of instances are selected into the training batch to form the contrastive loss, negative pairs that share similar semantics will be undesirably pushed apart in the embedding space.

\begin{figure}
\centering
\subfigure[Performance comparison on CUB]{
\label{fig:cubdimension}
\includegraphics[height=0.3\textwidth]{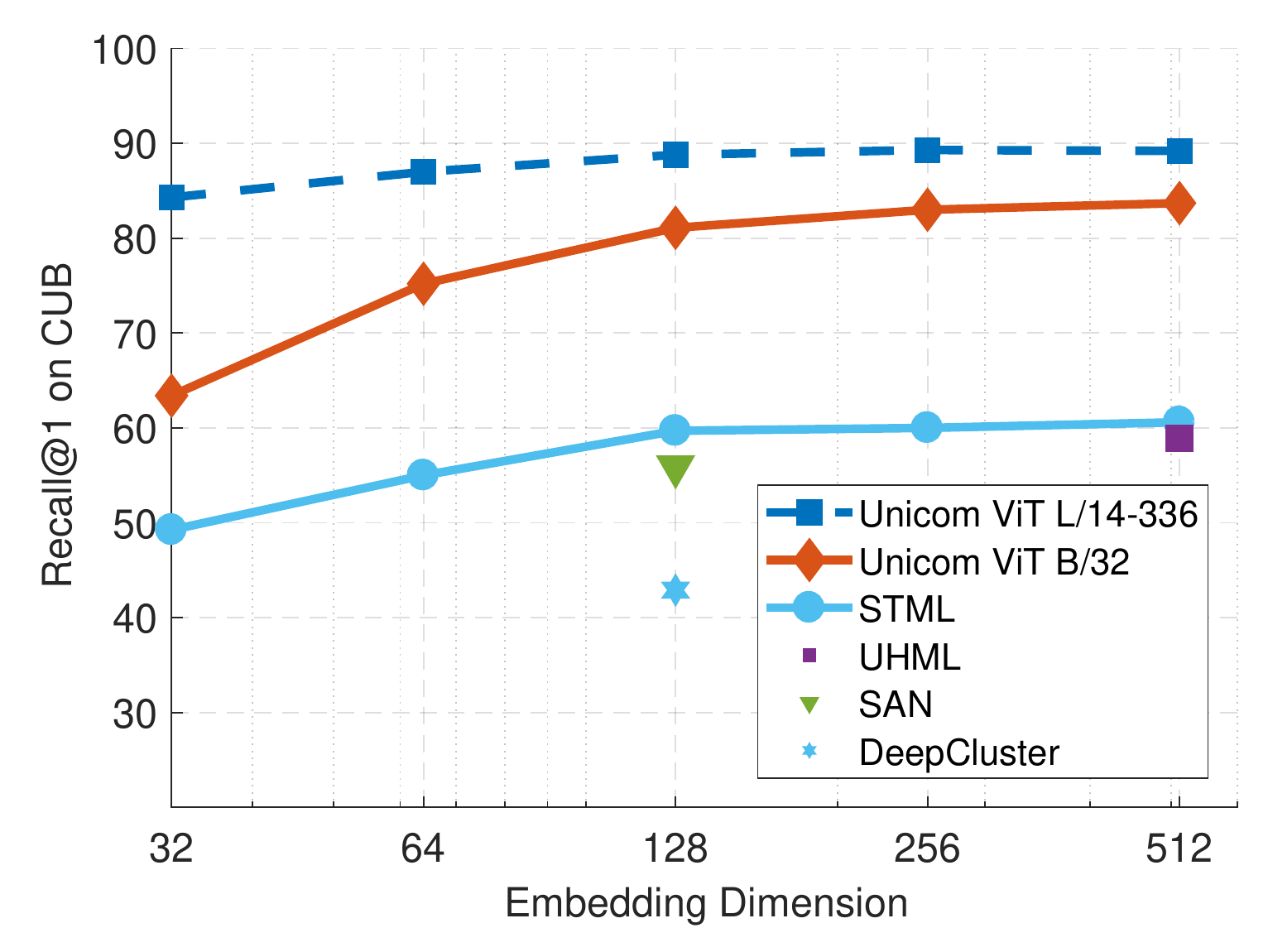}}
\subfigure[The proposed Unicom]{
\label{fig:2dpartialfc}
\includegraphics[height=0.3\textwidth]{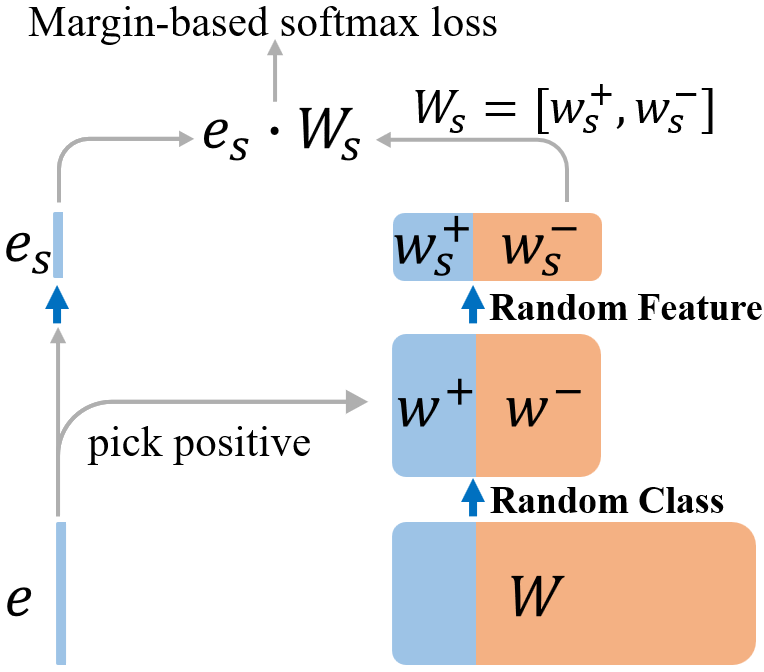}}
\caption{(a) Accuracy in Recall@1 versus embedding dimension on
the CUB dataset. The proposed Unicom is only trained on the \textcolor{black}{\laion{}} without any manual annotation. (b) The proposed Unicom employs two random selections along the class dimension and the feature dimension to alleviate inter-class conflict and achieve compact representation, respectively.}
\vspace{-4mm}
\label{fig:performanceandmethod}
\end{figure}

To handle the limitations of instance discrimination, cluster discrimination has been proposed for deep unsupervised learning through jointly learning image embeddings and cluster assignments. 
Learning representations with clusters will pull similar instances together, which is beneficial for capturing semantic structures in data. DeepCluster \citep{caron2018deep} performs iterative clustering by k-means and classification by cross-entropy loss, while SeLa \citep{asano2019self} proposes to solve an optimal transport problem for balanced assignment. However, both DeepCluster and SeLa need labels to be assigned offline in a batch mode with representations of all instances. To reduce the cost of batch mode clustering, ODC \citep{zhan2020online}, SwAV \citep{caron2020unsupervised}, and CoKe \citep{qian2022unsupervised} apply online clustering to avoid the multiple iterations over the entire dataset. 
Despite improved efficiency, the online clustering method still suffers from the collapsing problem (i.e., a dominant cluster includes instances from multiple classes or most of the instances).

\textcolor{black}{In this paper, we aim at boosting the semantic embedding power of the CLIP model by introducing a novel cluster discrimination approach.}
We first conduct one step of off-line clustering by using the image and text features from a pre-trained CLIP model \citep{radford2021learning}. Due to the limited discrimination power of the CLIP model, the pseudo clusters contain heavy inter-class conflict. 
Instead of optimizing the cluster assignment \citep{qian2022unsupervised}, we focus on how to robustly train a classifier on the automatically clustered large-scale data. 
More specifically, we explore two random selections on the prototype matrix $W\in {R}^{d\times k}$ when preparing the classification loss (as illustrated in Fig.~\ref{fig:2dpartialfc}). The first one is with respect to the class dimension ($k$), and only part of negative prototypes are selected for inter-class comparisons, which helps alleviate inter-class conflict. 
The second one is with respect to the feature dimension ($d$), and only part of features are randomly selected to construct the classification loss, enhancing the representation power of each neuron and making feature representation compact for the efficient image retrieval task. Concurrent with our work, partial selection mechanisms along class and feature are separately proposed in \citep{an2021partial,an2022killing} and \citep{xu2022subface} to accelerate model training and enhance locally distinguishable features. Both of their experiments are conducted on cleaned face recognition datasets. By contrast, we target at universal and compact representation learning from automatically clustered large-scale data. The main contributions of our paper are the following:
\vspace{-2mm}
\begin{tight_itemize}
\item We propose a novel cluster discrimination method for universal and compact representation learning. In the clustering step, we employ both image and text features from the pre-trained CLIP model. In the discrimination step, we explore two random selections along class and feature, which can potentially alleviate inter-class conflict and improve the feature compactness, respectively.
\item \textcolor{black}{For both zero-shot learning tasks (e.g., linear probe and unsupervised image retrieval) and transfer learning tasks (e.g., ImageNet-1K classification and supervised image retrieval),} the proposed random negative prototype selection for conflict-robust cluster discrimination can significantly boost the representation power compared to the instance discrimination based model (e.g., CLIP).
\end{tight_itemize}

\section{Related Work}
\noindent{\bf Visual Model Pre-training.} Model pre-training for visual recognition can be categorized into three main groups: (1) supervised pre-training on datasets with manually annotated class labels (e.g., ImageNet-1K/-21K \citep{deng2009imagenet} and JFT-300M/-3B \citep{dosovitskiy2021image,zhai2022scaling}), (2) weakly-supervised pre-training by using hashtag \citep{mahajan2018exploring,singh2022revisiting} or \textcolor{black}{text descriptions \citep{radford2021learning,jia2021scaling}}, and (3) unsupervised pre-training \citep{chen2020big,he2020momentum,caron2018deep}. Since supervised pre-training relies on expensive manual annotations, we focus on \textcolor{black}{annotation-free pre-training} which has the advantages of being easily scaled to billions of training images and being able to learn universal feature representations for downstream tasks. 

\noindent{\bf Instance and Cluster Discrimination.} Instance discrimination \citep{chen2020big,he2020momentum,radford2021learning} is realized with a contrastive loss which targets at pulling closer samples from the same instance while pushing away samples from different instances. Despite the impressive performance, instance-wise contrastive learning can not capture the semantic information from the training data because it is trained to ignore the similarity between different instances. Cluster discrimination \citep{caron2018deep,zhan2020online,li2020prototypical} is processed with iterative steps: the clustering step to assign pseudo class labels for each sample, and then the classification step to map each sample to its assigned label. 
Since one cluster has more than one instance, learning representations with clusters will gather similar instances together, which can explore potential semantic structures in data. As a representative work, DeepCluster \citep{caron2018deep} adopts a standard k-means for clustering, but it contains degenerate solutions. To this end, recent research work \citep{asano2019self,caron2020unsupervised,qian2022unsupervised} focuses on improving the label assignment during clustering but employs a standard cross-entropy loss during discrimination. 
In this paper, we only employ one step of off-line \textcolor{black}{clustering} but design a robust classifier to achieve good feature representation when training on the automatically clustered large-scale data.

\noindent{\bf Image Retrieval.} Image retrieval task typically relies on fine-tuning pre-trained visual models \citep{szegedy2015going,he2016deep,dosovitskiy2021image} and can be divided into two learning categories: supervised and unsupervised metric learning. For supervised metric learning, pair-wise loss \citep{hadsell2006dimensionality,schroff2015facenet,sohn2016improved} and cross-entropy loss \citep{zhai2018classification,deng2019arcface,sun2020circle,qian2019softtriple} are extensively studied and recent bench-marking results \citep{musgrave2020metric} indicate that the margin-based softmax loss (e.g., ArcFace~\citep{deng2019arcface}) can achieve state-of-the-art performance. For unsupervised metric learning, pseudo labeling methods are employed to discover pseudo classes by applying k-means clustering \citep{kan2021relative,li2020unsupervised}, hierarchical clustering \citep{yan2021unsupervised}, random walk \citep{iscen2018mining}, and class-equivalence relations \citep{Kim2022CVPR} to unlabeled training data. 
In this paper, we focus on universal and compact feature embedding for both unsupervised and supervised image retrieval task.

\section{Methodology}
\subsection{Preliminaries of Instance and Cluster Discrimination}
Given a training set $X=\{x_1, x_2,...,x_n\}$ including $n$ images,
feature representation learning aims at learning a function $f$ that maps \textcolor{black}{images $X$} to embeddings $E=\{e_1, e_2,...,e_n\}$ with $e_i=f(x_i)$, such that embeddings can describe the similarities between different images. 
Instance discrimination achieves this objective by optimizing a contrastive loss function defined as:
\vspace{-2mm}
\begin{equation}
\label{eqn:ID}
\mathcal{L}_\mathrm{instance} = - \sum_{i=1}^n  \log \frac{\exp(e_i'^T e_i   )}{\sum_{j=0}^m \exp(e_j'^T e_i   ) },
\end{equation}
where $e_i$ and $e_i'$ are positive embeddings of the instance $i$, and $e_j'$ consists of one positive embedding of $i$ and its $m$ negative embeddings from other instances.

By contrast, cluster discrimination for representation learning consists of two main phases: clustering and discrimination. 
The clustering step assigns each instance a pseudo class label that will be subsequently used as supervision to train a classifier in the discrimination step. 
Following this, automatic clustering on the features $e_i = f(x_i)$ is first performed 
to obtain $k$ clusters and the centroid $w_i$ is viewed as the prototype of $i$-th cluster. 
Then, the training data $\{x_i\}_{i=1}^n$ are  partitioned into $k$ classes represented by prototypes $W=\{w_i\}_{i=1}^k$. 
With pseudo labels and centroids obtained from the clustering step, cluster discrimination can be implemented by optimizing a standard softmax classification loss as:
\vspace{-2mm}
\begin{equation}
\label{eqn:pl}
\mathcal{L}_\mathrm{cluster} = - \sum_{i=1}^n  \log \frac{\exp( w_i^T e_i)}{\sum_{j=1}^{k} \exp(w_j^Te_i)}, 
\end{equation}
where $e_i$ is the embedding of the image $x_i$ and $x_i$ belongs to the class represented by $w_i$. 
By comparing Eq.~\ref{eqn:ID} and Eq.~\ref{eqn:pl}, we can observe the difference  that instance discrimination employs an augmented feature $e_i'$ to calculate the similarities while cluster discrimination uses a prototype $w_i$.

\begin{figure}
\centering
\subfigure[Multi-modal Clustering]{
\label{fig:multimodalclustering}
\includegraphics[height=0.32\textwidth]{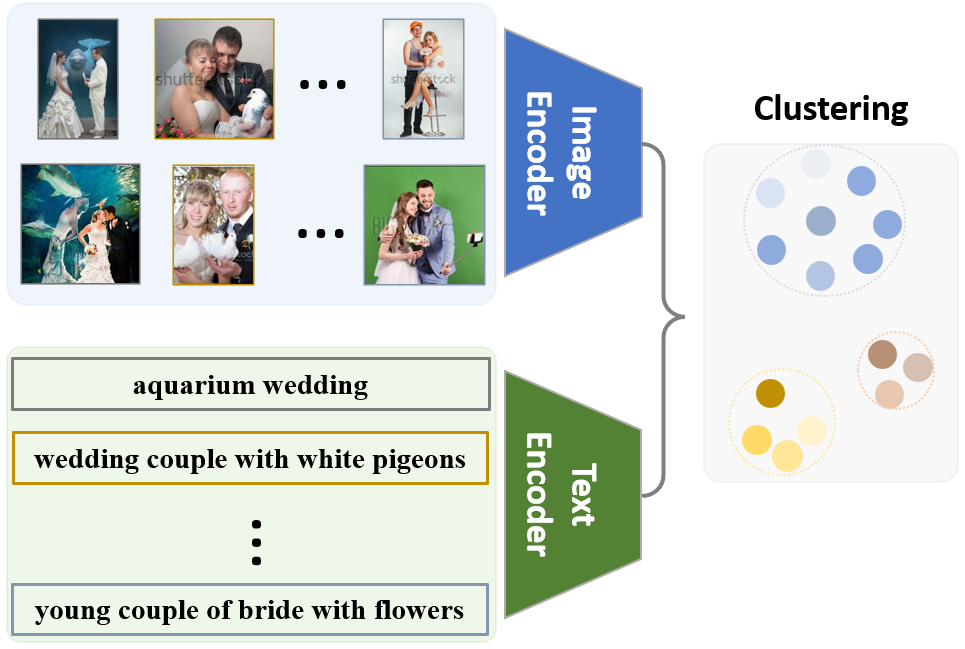}}
\hfill
\subfigure[Discrimination]{
\label{fig:discrimination}
\includegraphics[height=0.32\textwidth]{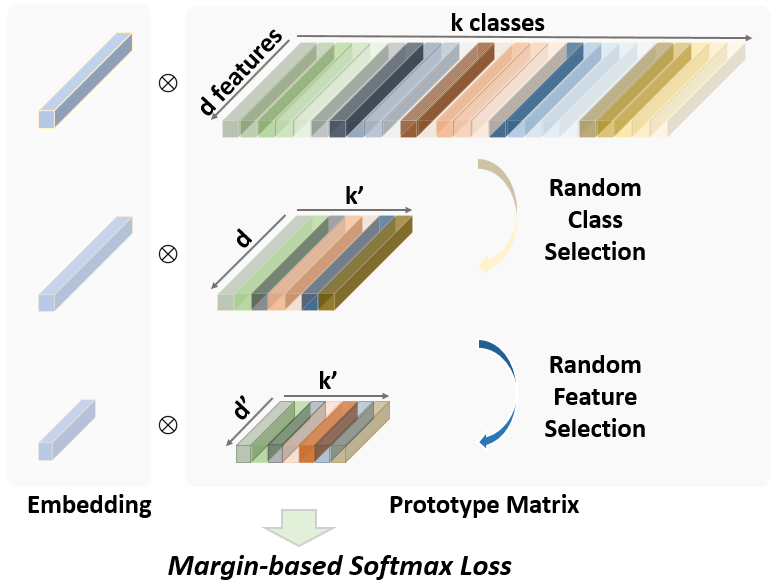}}
\caption{Illustration of the proposed method. (a) The multi-modal clustering includes one off-line step of k-means on features from image and text produced by a pre-trained CLIP model \citep{radford2021learning}. (b) Using the assigned clusters as pseudo-labels, we propose a conflict-robust and representation-compact classification method through random class and feature selection along the two dimensions of the prototype matrix.}
\vspace{-4mm}
\label{fig:clusterdiscrimination}
\end{figure}

\subsection{Multimodal Clustering} \label{sec:mcluster}
In this paper, we focus on the standard clustering algorithm, $k$-means, which takes a set of vectors as input and clusters them into $k$ distinct groups based on the nearest neighbor criterion. 
To seek a better representation, we combined the image and text features produced by the pre-trained CLIP model \citep{radford2021learning} due to their mutual complementary nature. 
The clustering step jointly learns a $d\times k$ centroid matrix $W$ and the cluster assignments $y_i$ of each image $x_i$ by solving the following problem:
\vspace{-2mm}
\begin{equation}
\label{eq:kmeans}
\min_{W \in \mathbb{R}^{d\times k}}
\frac{1}{n}
\sum_{i=1}^n
\min_{y_i \in \{0,1\}^{k}}
\| \Phi (f(x_i), f'(x'_i))  -  W y_i \|_2^2
\quad
\text{s.t.}
\quad
y_i^\top \bf{1}_k = 1,
\vspace{-2mm}
\end{equation}
where $f(x_i)$ is the image feature embedding by the image encoder $f$ and $f'(x'_i)$ is the text feature embedding by the text encoder $f'$, $\Phi$ is a feature ensemble function, $W\in {R}^{d\times k}$ is the centroid matrix, $y_i$ in $\{0,1\}^k$ is a single label assignment constrained by $y_i^\top \bf{1}_k = 1$, and $ \bf{1}_k $ is 1-vector of size $k$. In this work, we employ the simplest feature ensemble function, that is, averaging the image and text features, as the CLIP model provides an aligned visual-textual representation.

\begin{figure}
\centering

\subfigure[Wedding+Flower]{
\label{fig1:wedding1}
\includegraphics[height=0.22\textwidth]{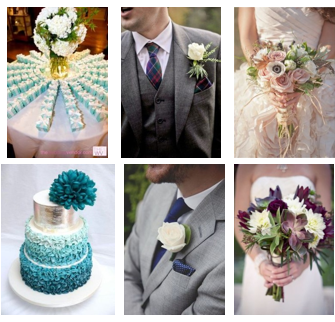}}
\subfigure[Wedding+Pair]{
\label{fig1:wedding2}
\includegraphics[height=0.22\textwidth]{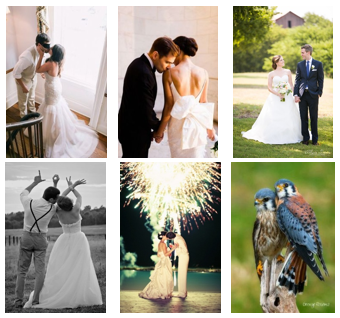}}
\subfigure[Wedding+Place]{
\label{fig1:wedding3}
\includegraphics[height=0.22\textwidth]{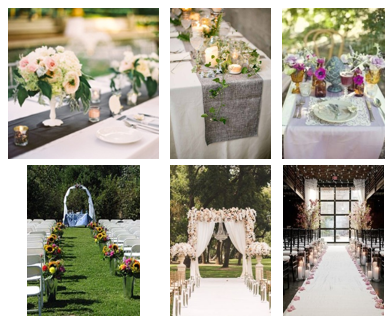}}
\caption{\textcolor{black}{Inter-class conflict between the automatically-clustered classes}. The class name is given based on the observation of images and texts. Inter-class conflict exists due to specific granularity definitions and multi-label signals in one image.}
\vspace{-4mm}
\label{fig1:clutering}
\end{figure}

Considering that iterative clustering and discrimination are time-consuming, we only employ one step of off-line clustering in this work. Aided by the efficient feature quantization \citep{johnson2019billion}, the large-scale \laion{} can be clustered within 10 minutes using the embedded image and text features. 
The only hyper-parameter we consider here is the cluster number $k$. Even though the clustering step is straightforward, the automatically clustered large-scale dataset inevitably confronts inter-class conflict due to multi-label signals in one image and specific definition of class granularity (as illustrated in Fig.~\ref{fig1:clutering}). 
For instance, the bird pair in Fig.~\ref{fig1:wedding2} is clustered into wedding pair, which will be conflicted with another specific category of bird. In addition, the close-up capture of wedding flowers in Fig.~\ref{fig1:wedding1} also exists in the class of wedding place, where flowers are the most popular decoration.

\subsection{conflict-robust and Representation-compact Discrimination}
After the clustering step, we can employ a vanilla classification loss (e.g., softmax) to learn the feature representation. 
For the softmax loss given in Eq.~\ref{eqn:pl}, the derivatives to a class-wise prototype ${w}_j\in\mathbb{R}^d$ and a sample embedding feature ${e}_{i}\in\mathbb{R}^d$ are:
\vspace{-2mm}
\begin{equation}
\frac{\partial \mathcal{L}}{\partial w_j} = \sum_{i=1}^{b}(p_{ij}-\mathbbm{1} \{y_i==j\}){e}_i,~~~~~~
\frac{\partial \mathcal{L}}{\partial {e}_{i}}=\sum_{j=1}^{k}(p_{ij}-\mathbbm{1} \{y_i==j\} ){w}_{j},  
\label{PL-derivative} 
\vspace{-4mm}
\end{equation}
where $b$ is the batch size, $k$ is the class number, and $p_{ij} = {e^{{w}_{j}^T{e}_{i}}}/{\sum_{l=1}^{k}e^{{w}_{l}^{{T}}{e}_{i}}}$ is the normalized probability of the sample embedding ${e}_{i}$ belonging to the prototype ${w}_{j}$, $\mathbbm{1} (\cdot)$ is the indicator function which is $1$ when the statement is true and $0$ otherwise. 
In Eq.~\ref{PL-derivative}, the derivative of the prototype is a ``weighted sum'' over sample features from the mini-batch, and the derivative of the sample feature is a ``weighted sum'' over the prototypes of all classes. 
If conflicted classes exist, wrong gradient signals from these conflicted negative prototypes will affect the update of model parameters. 

To this end, we propose a random negative prototype selection to efficiently construct a negative prototype subset from the entire negative prototypes. \textcolor{black}{Therefore, the derivative to a sample embedding feature is:
\vspace{-2mm}
\begin{gather}
\frac{\partial \mathcal{L}}{\partial {e}_{i}} = -((1-p^+)w^+ - \sum_{j\in \mathbb{S},j\ne y_i}   p^-_j w^-_j),  
\label{x-gradientpfc} 
\vspace{-6mm}
\end{gather}
where $p^+$ and $w^+$ denote the probability and prototype of the positive class, $p^-_j$ and $w^-_j$ refer to negative probabilities and prototypes, $\mathbb{S}$ is a subset of all negative classes and one positive class, $\left |\mathbb{S}  \right |= k*r_1$, and $r_1 \in\left [0,1 \right ]$ is the sampling ratio.}
Even though all class-wise prototypes are still maintained throughout the whole training process, only positive prototypes and a subset of negative prototypes are selected and updated in each iteration.
Therefore, the inter-class conflict will be reduced as the possibility of sampling a conflict negative prototype is directly decreased by $r_1$.

To achieve a compact representation for efficient image retrieval, previous methods \citep{babenko2015aggregating,tolias2016particular} adopt Principal Component Analysis (PCA) on an independent set for dimension reduction. 
To reduce the descriptor dimension to $d'$, only eigenvectors corresponding to $d'$ largest eigenvalues are retained. 
However, the representation produced by PCA is sub-optimal because it is a post-processing step detached from the target task.
To this end, we propose a feature approximation strategy by randomly selecting subspace features to construct the classification loss: 

\vspace{-2mm}
\begin{equation}
\label{eqn:unicom}
\mathcal{L}_\mathrm{unicom} = - \sum_{i=1}^n  \log \frac{\exp( (\Gamma_t \odot w_i)^T (\Gamma_t \odot e_i))}{\sum_{j\in \mathbb{S}} \exp( (\Gamma_t \odot w_j)^T (\Gamma_t \odot e_i))}, 
\end{equation}
where $\Gamma_t \in \{0,1\} ^d$ is a random binary vector at the iteration step of $t$, \textcolor{black}{the non-zero element ratio of $\Gamma_t$ is $r_2 \in\left [0,1 \right ]$}, $\odot$ denotes element-wise product.
\textcolor{black}{
Different from the well-known regularization technique, Dropout~\citep{srivastava2014dropout}, our random feature selection $\Gamma_t$ is same for all training samples within the mini-batch. $\Gamma_t$ is applied to both feature $e_i$ and prototypes $w_j$, thus the dimension of derivatives in Eq.~\ref{PL-derivative} decreases to $d'$. By contrast, Dropout is independently applied to each individual feature $e_i$ within the mini-batch by setting a specific ratio $r_3\in\left [0,1 \right ]$ of neurons to $0$ and enlarging the rest neurons by $1/(1-r_3)$. The dimension of derivatives in Eq.~\ref{PL-derivative} is still $d$. Therefore, the sub-feature optimization in the proposed random feature selection can not be completed by directly calling the Dropout function.} 
Since the binary vector varies at different iterations, different sub-features are trained and sub-gradients are calculated. This leads to a solution that each embedding neuron contains the similarity representation power.

The schematic of the proposed method is in Fig.~\ref{fig:discrimination}.
As shown, the prototype matrix $W$ is maintained in the memory at the dimension of $d \times k$ during the whole training process, but \textcolor{black}{only part of the classes ($k'=k*r_1$) and features ($d'=d*r_2$) are randomly selected to construct the softmax loss}. 
The first random selection along the class dimension is for conflict-robust learning to achieve universal representation and the second random selection along the feature dimension is for feature compression required by efficient retrieval. Therefore, we name our method UNIversal and COMpact (UNICOM) representation learning.

\section{Experiments}
\subsection{Implementation Details}
Unless otherwise specified, all ViT models in our experiments follow the same architecture designs in CLIP, and are \textcolor{black}{trained from scratch} for 32 epochs on the automatically clustered \laion{} (Section \ref{sec:mcluster}) with cluster number $k=1M$.
During training, we randomly crop and horizontally flip each image to get the input image with $224\times224$ resolution. 
We set the random class sampling ratio $r_1$ as $0.1$ in the pre-training step.
The training is conducted on 128 NVIDIA V100 GPUs across 16 nodes. 
To save memory and scale up the batch size, mixed-precision and gradient checkpoint are used. We use AdamW \citep{loshchilov2018decoupled} as the optimizer with an initial learning rate of $0.001$, and a weight decay of $0.05$.
We employ margin-based softmax loss, ArcFace \citep{deng2019arcface,deng2020sub}, for both pre-training and image retrieval tasks. 
The margin value is set to $0.3$ and the feature scale is set to $64$.
For supervised retrieval, we follow the data-split settings of the baseline methods \citep{Patel2022CVPR,Ermolov_2022_CVPR} to fine-tune models.

\subsection{Comparisons on Feature Representation Learning}
\textcolor{black}{In this section, we first compare the performance of the proposed method and other baseline models (i.e., CLIP and OPEN-CLIP) on the linear probe and unsupervised image retrieval. Specifically, after the training on the automatically clustered 1M classes, we fix the backbones of our models. For the linear probe task, we learn an additional FC layer for classification on each test set. For unsupervised image retrieval, we directly use the embedding features for testing. Then, we fine-tune the pre-trained models for supervised image retrieval on each image retrieval dataset.}

\noindent{\bf Linear Probe.} Following the same evaluation setting as CLIP \citep{radford2021learning}, we freeze the pre-trained models on \laion{} and only fine-tune the last linear classification layer. 
We report the linear probing performance over 13 datasets in Tab.~\ref{tab:linear-probe-big-table}. 
The proposed conflict-robust cluster discrimination method significantly outperforms the CLIP and OPEN-CLIP \citep{openclip} models. 
Notably, our ViT B/32, ViT B/16, and ViT L/14 models surpass counterparts of OPEN-CLIP by 3.6\%, 2.7\% and 1.4\% on average with the same training data, indicating that the proposed cluster discrimination can enhance the representation power over instance discrimination.

\noindent{\bf Unsupervised Image Retrieval.}
In Tab.~\ref{tab:zeroshotimageretrieval}, we compare the performance of unsupervised image retrieval by directly using the pre-trained models for feature embedding. The GLDv2 \citep{weyand2020google} employs mean Average Precision@100 (mAP@100) as the evaluation metric, while other datasets use Recall@1.
Our ViT L/14 model achieves 69.9\% average result across 7 image retrieval datasets, surpassing the OPEN-CLIP counterpart by 7.5\% and even outperforming the larger OPEN-CLIP model ViT H/14 by $5.4\%$. 
The reason behind this significant improvement is that the proposed cluster discrimination can capture the semantic structure in data, which is crucial for the image retrieval task. 
In Fig.~\ref{fig:cubdimension}, we compare our method with the state-of-the-art unsupervised image retrieval approach, STML~\citep{Kim2022CVPR}, under different dimension constraints on the CUB dataset. We set the random feature selection ratio $r_2$ as $0.5$ for one additional training epoch on the \laion{}. Then, we select the first 256-D, 128-D, 64-D, and 32-D features for testing. STML employs an ImageNet-1K pre-trained GoogleNet \citep{szegedy2015going} and then explores unsupervised training on the CUB dataset. Even though our ViT-based model is only trained on the automatically clustered \laion{} without any further training on the image retrieval dataset, our method outperforms STML~\citep{Kim2022CVPR} by a large margin across different test dimensions, indicating the superiority of the proposed random feature selection for compact feature representation learning.

\begin{table}
	\caption{Top-1 accuracy(\%) of linear probe on 13 image classification datasets. The proposed cluster discrimination significantly outperforms OPEN-CLIP \citep{openclip} on average by using the same training data (i.e., LAION 400M). ``CLIP-R'' denotes testing the public CLIP-ViT models in our code base. ``-336'' refers to one additional epoch of pre-training at a higher $336\times336$ resolution to boost performance.}
	\label{tab:linear-probe-big-table}
	\centering
	\resizebox{1.0\linewidth}{!}{
		\begin{tabular}{cl|cccccccccccccc|c}
			\toprule
			 &              &  & \rotatebox[origin=lb]{90}{\smash{CIFAR10}} & \rotatebox[origin=lb]{90}{\smash{CIFAR100}} & \rotatebox[origin=lb]{90}{\smash{Caltech101}} & \rotatebox[origin=lb]{90}{\smash{Cars}} & \rotatebox[origin=lb]{90}{\smash{Flowers}} & \rotatebox[origin=lb]{90}{\smash{Food101}} & \rotatebox[origin=lb]{90}{\smash{Birdsnap}} & \rotatebox[origin=lb]{90}{\smash{SUN397}} & \rotatebox[origin=lb]{90}{\smash{DTD}} & \rotatebox[origin=lb]{90}{\smash{Aircraft}} & \rotatebox[origin=lb]{90}{\smash{Pets}} & \rotatebox[origin=lb]{90}{\smash{EuroSAT}} & \rotatebox[origin=lb]{90}{\smash{ImageNet}} & \rotatebox[origin=lb]{90}{\smash{Average}} \\
			\midrule
			\multirow{4}{0em}{\rotatebox[origin=c]{90}{CLIP}}
			 & ViT B/32     &  & 95.1                                       & 80.5                                        & 93.0                                          & 81.8                                    & 96.9                                       & 88.8                                       & 58.5                                        & 76.6                                      & 76.5                                   & 52.0                                        & 90.0                                    & 97.0                                       & 76.1                                        & 81.8                                        \\
			 & ViT B/16     &  & 96.2                                       & 83.1                                        & 94.7                                          & 86.7                                    & 98.1                                       & 92.8                                       & 67.8                                        & 78.4                                      & 79.2                                   & 59.5                                        & 93.1                                    & 97.1                                       & 80.2                                        & 85.1                                        \\
			 & ViT L/14     &  & 98.0                                       & 87.5                                        & 96.5                                          & 90.9                                    & 99.2                                       & 95.2                                       & 77.0                                        & 81.8                                      & 82.1                                   & 69.4                                        & 95.1                                    & 98.2                                       & 83.9                                        & 88.8                                        \\
			 & ViT L/14-336 &  & 97.9                                       & 87.4                                        & 96.0                                          & 91.5                                    & 99.2                                       & 95.9                                       & 79.9                                        & 82.2                                      & 83.0                                   & 71.6                                        & 95.1                                    & 98.1                                       & 85.4                                        & 89.5                                        \\ \midrule \multirow{4}{0em}{\rotatebox[origin=c]{90}{CLIP-R}}
			 & ViT B/32     &  & 96.0                                       & 82.5                                        & 94.1                                          & 86.0                                    & 97.8                                       & 92.7                                       & 61.1                                        & 79.1                                      & 78.4                                   & 58.9                                        & 93.0                                    & 95.3                                       & 75.3                                        & 83.9                                        \\
			 & ViT B/16     &  & 96.0                                       & 82.5                                        & 94.1                                          & 86.0                                    & 97.8                                       & 92.7                                       & 69.5                                        & 79.1                                      & 78.4                                   & 58.9                                        & 93.0                                    & 95.3                                       & 79.6                                        & 84.8                                        \\
			 & ViT L/14     &  & 98.1                                       & 87.2                                        & 96.0                                          & 90.7                                    & 99.2                                       & 95.3                                       & 77.8                                        & 81.5                                      & 80.9                                   & 68.0                                        & 94.9                                    & 96.7                                       & 84.1                                        & 88.5                                        \\
			 & ViT L/14-336 &  & 97.8                                       & 87.1                                        & 96.3                                          & 91.4                                    & 99.2                                       & 95.9                                       & 80.9                                        & 82.2                                      & 82.4                                   & 71.2                                        & 95.1                                    & 96.8                                       & 84.9                                        & 89.3                                        \\ \midrule \multirow{3}{0em}{\rotatebox[origin=c]{90}{OPEN}}
			 & ViT B/32     &  & 95.3                                       & 82.2                                        & 93.3                                          & 87.5                                    & 96.5                                       & 86.2                                       & 61.4                                        & 75.3                                      & 78.8                                   & 52.4                                        & 88.0                                    & 96.5                                       & 73.8                                        & 82.1                                        \\
			 & ViT B/16     &  & 96.4                                       & 84.0                                        & 94.1                                          & 91.8                                    & 98.1                                       & 90.7                                       & 71.2                                        & 78.7                                      & 81.6                                   & 59.3                                        & 90.0                                    & 96.2                                       & 78.5                                        & 85.4                                        \\
			 & ViT L/14     &  & 97.9                                       & 87.9                                        & 95.5                                          & 93.6                                    & 98.8                                       & 93.3                                       & 78.0                                        & 81.0                                      & 83.0                                   & 64.4                                        & 93.3                                    & 97.1                                       & 81.5                                        & 88.1                                        \\ \midrule \multirow{4}{0em}{\rotatebox[origin=c]{90}{Ours}}
			 & ViT B/32     &  & 96.8                                       & 86.6                                        & 94.6                                          & 93.3                                    & 98.5                                       & 85.8                                       & 70.2                                        & 74.6                                      & 78.0                                   & 70.7                                        & 93.1                                    & 96.8                                       & 75.0                                        & 85.7                                        \\
			 & ViT B/16     &  & 97.3                                       & 87.7                                        & 95.1                                          & 94.3                                    & 98.9                                       & 91.2                                       & 79.3                                        & 77.1                                      & 81.2                                   & 73.4                                        & 93.9                                    & 97.0                                       & 79.1                                        & 88.1                                        \\
			 & ViT L/14     &  & 98.5                                       & 90.8                                        & 95.7                                          & 94.6                                    & 99.3                                       & 93.4                                       & 82.4                                        & 80.0                                      & 82.2                                   & 74.5                                        & 94.2                                    & 96.7                                       & 81.8                                        & 89.5                                        \\
			 & ViT L/14-336 &  & 98.5                                       & 90.7                                        & 95.7                                          & 95.1                                    & 99.4                                       & 94.3                                       & 85.1                                        & 79.7                                      & 82.0                                   & 78.1                                        & 94.5                                    & 97.2                                       & 82.7                                        & 90.2                                        \\
			\bottomrule
		\end{tabular}
	}
\end{table}

\begin{table}
	\caption{Performance of unsupervised image retrieval on 7 image retrieval datasets. The proposed conflict-robust cluster discrimination significantly outperforms OPEN-CLIP on average by using the same training data.}
	\label{tab:zeroshotimageretrieval}
	\resizebox{1.0\linewidth}{!}{
		\begin{tabular}{cl|ccccc|ccc|cc|c}
			\toprule
			 &          & CUB  & Cars & SOP  & In-Shop & INaturalist & \multicolumn{3}{c|}{VehicleID} & \multicolumn{2}{c|}{GLDv2} & Average                           \\
			 &          &      &      &      &         &             & Small                          & Medium                     & Large   & Private & Public &      \\
			\midrule \multirow{4}{01em}{\rotatebox[origin=c]{90}{CLIP}}
			 & B/32     & 56.7 & 79.0 & 60.5 & 45.4    & 53.0        & 54.8                           & 52.2                       & 44.6    & 7.5     & 7.5    & 46.1 \\
			 & B/16     & 66.1 & 85.2 & 63.2 & 56.1    & 63.1        & 55.1                           & 50.9                       & 43.8    & 8.4     & 10.6   & 50.3 \\
			 & L/14     & 76.0 & 90.3 & 65.6 & 62.9    & 72.9        & 62.4                           & 58.9                       & 51.8    & 12.1    & 13.6   & 56.7 \\
			 & L/14-336 & 77.3 & 90.9 & 67.8 & 66.3    & 76.8        & 64.1                           & 60.3                       & 53.8    & 17.0    & 15.6   & 59.0 \\ \midrule \multirow{4}{01em}{\rotatebox[origin=c]{90}{OPEN}}
			 & B/32     & 62.3 & 89.2 & 65.9 & 64.6    & 54.9        & 71.0                           & 67.1                       & 59.9    & 9.17    & 8.4    & 55.2 \\
			 & B/16     & 71.4 & 92.9 & 68.7 & 74.2    & 64.1        & 73.3                           & 70.1                       & 63.7    & 12.1    & 11.0   & 60.2 \\
			 & L/14     & 79.4 & 94.9 & 70.6 & 77.1    & 71.0        & 72.0                           & 69.1                       & 62.0    & 14.5    & 13.8   & 62.4 \\
			 & H/14     & 83.1 & 95.7 & 72.7 & 78.8    & 77.0        & 72.7                           & 69.7                       & 61.9    & 17.7    & 15.3   & 64.5 \\ \midrule \multirow{4}{01em}{\rotatebox[origin=c]{90}{Ours}}
			 & B/32     & 83.7 & 95.9 & 70.0 & 72.8    & 64.6        & 74.9                           & 72.0                       & 65.4    & 15.1    & 13.3   & 62.8 \\
			 & B/16     & 86.5 & 96.8 & 70.4 & 74.6    & 73.6        & 74.5                           & 70.6                       & 58.7    & 18.7    & 17.2   & 64.2 \\
			 & L/14     & 88.5 & 96.9 & 72.7 & 83.6    & 77.1        & 83.7                           & 80.2                       & 74.6    & 21.1    & 20.1   & 69.9 \\
			 & L/14-336 & 89.2 & 97.3 & 74.5 & 86.7    & 81.0        & 84.1                           & 81.4                       & 75.6    & 23.2    & 22.0   & 71.5 \\
			\bottomrule
		\end{tabular}
	}
\end{table}

\begin{table}[t]
	\caption{Transfer-learning accuracy of models pre-trained on the specified dataset followed by fine-tuning and testing on ImageNet.}
	\label{tab:imagenet_results}
	\centering
	\resizebox{0.9\columnwidth}{!}{
		\begin{tabular}{lccc|c|l}
			\toprule
			Model                                 & Pre-training & \multicolumn{2}{c|}{  Resolution} & IN-1K    & {FLOPs}                \\
			                                      & Dataset      & Pretrain                          & Finetune & Top-1 Accuracy & (B)   \\\midrule
			\multicolumn{6}{l}{\textit{Supervised pre-training}}                                                                         \\\midrule
			ViT L/32~\citep{dosovitskiy2021image} & IN-21k       & 224                               & 384      & 81.3           & 54.4  \\
			ViT B/16~\citep{dosovitskiy2021image} & IN-21k       & 224                               & 384      & 84.0           & 55.6  \\
			ViT L/16~\citep{dosovitskiy2021image} & IN-21k       & 224                               & 384      & 85.2           & 191.5 \\
			ViT L/16~\citep{dosovitskiy2021image} & JFT 300M     & 224                               & 512      & 87.8           & 362.9 \\
			ViT L/16~\citep{zhai2022scaling}      & JFT 3B       & 224                               & 384      & 88.5           & 191.5 \\
			\midrule
			\multicolumn{6}{l}{\textit{Weakly supervised pre-training}}                                                                  \\\midrule
			ViT B/16~\citep{singh2022revisiting}  & IG 3.6B      & 224                               & 384      & 85.3           & 55.6  \\
			ViT L/16~\citep{singh2022revisiting}  & IG 3.6B      & 224                               & 512      & 88.1           & 362.9 \\
			\midrule
			{ViT B/32 OPEN-CLIP}                  & LAION 400M   & 224                               & 384      & 83.0           & 15.5  \\
			{ViT B/16 OPEN-CLIP}                  & LAION 400M   & 224                               & 384      & 85.4           & 55.6  \\
			{ViT L/14 OPEN-CLIP}                  & LAION 400M   & 224                               & 518      & 87.7           & 507.8 \\
			{ViT B/32 Ours}                       & LAION 400M   & 224                               & 384      & 83.6           & 15.5  \\
			{ViT B/16 Ours}                       & LAION 400M   & 224                               & 384      & 85.9           & 55.6  \\
			{ViT L/14 Ours}                       & LAION 400M   & 224                               & 518      & 88.3           & 507.8 \\
			\bottomrule
		\end{tabular}
	}
\end{table}

\begin{table}
	\caption {\textcolor{black}{Performance of supervised image retrieval on 7 image retrieval datasets.}}
	\label{tbl:sota}
	\centering
	\resizebox{1.0\linewidth}{!}{
		\begin{tabular}	{l |c |c |c  |c | l}
			\toprule
			                 & ViT-B/32 & ViT-B/16 & ViT-L/14 & ViT-L/14-336 & \hspace{0.3cm} Previous SOTA                                      \\
			\midrule
			CUB              & 85.8     & 88.8     & 89.7     & 90.1         & \hspace{0.3cm} 85.6 {\scriptsize ViT-S/16}     \citep{Ermolov_2022_CVPR}  \\
			Cars             & 97.3     & 97.7     & 97.9     & 98.2         & \hspace{0.3cm} 94.8 {\scriptsize SE-ResNet-50} \citep{jun2019combination} \\
			SOP              & 87.1     & 88.8     & 89.9     & 91.2         & \hspace{0.3cm} 88.0 {\scriptsize ViT-B/16}     \citep{Patel2022CVPR}      \\
			In-Shop          & 94.8     & 95.5     & 96.0     & 96.7         & \hspace{0.3cm} 92.7 {\scriptsize ViT-S/16}     \citep{Ermolov_2022_CVPR}  \\
			INaturalist      & 72.8     & 82.5     & 85.4     & 88.9         & \hspace{0.3cm} 83.9 {\scriptsize ViT-B/16}     \citep{Patel2022CVPR}      \\
			VehicleID-Small  & 95.4     & 96.4     & 96.5     & 97.0         & \hspace{0.3cm} 96.2 {\scriptsize ViT-B/16}     \citep{Patel2022CVPR}      \\
			VehicleID-Medium & 94.1     & 95.1     & 95.7     & 96.1         & \hspace{0.3cm} 95.2 {\scriptsize ViT-B/16}     \citep{Patel2022CVPR}      \\
			VehicleID-Large  & 93.6     & 95.0     & 95.4     & 96.0         & \hspace{0.3cm} 94.7 {\scriptsize ViT-B/16}     \citep{Patel2022CVPR}      \\
			GLDv2-Private    & 32.6     & 35.7     & 36.1     & 36.4         & \hspace{0.3cm} 32.5 {\scriptsize ResNet101}    \citep{Lee_2022_CVPR}      \\
			GLDv2-Public     & 29.7     & 32.4     & 33.0     & 33.1         & \hspace{0.3cm} 24.6 {\scriptsize ResNet50}     \citep{Tan_2021_ICCV}      \\
			\bottomrule
		\end{tabular}}
\end{table}

\begin{table}[t]
	\caption {\textcolor{black}{Ablation study on multi-modal clustering.} ViT B/32 is used here for model training on the LAION 400M dataset, which is automatically clustered by different pre-trained models. We report the average performance of linear probe and unsupervised image retrieval.}
	\label{tbl:abalationclusterand1strandom}
	\centering
	\resizebox{0.8\linewidth}{!}{
		\begin{tabular}	{l  |  c  c  c | c  c  c| c c  c  }
			\toprule
			Tasks        & \multicolumn{3}{c|}{CLIP} & \multicolumn{3}{c|}{OPEN-CLIP} & \multicolumn{3}{c}{Cluster Number by CLIP}                                             \\
			             & Image                     & Text                           & Joint                                      & Image & Text & Joint & 100K & 1M   & 10M  \\ \midrule
			Linear Probe & 84.4                      & 85.3                           & 85.7                                       & 83.9  & 84.0 & 84.1  & 75.9 & 85.7 & 83.6 \\
			Unsup. Retr. & 61.8                      & 62.3                           & 62.8                                       & 58.9  & 60.1 & 61.1  & 53.2 & 62.8 & 60.7 \\
			\bottomrule
		\end{tabular}
	}
\end{table}

\begin{table}[t]
	\caption {\textcolor{black}{Ablation study on random negative class selection and random feature selection.} ViT-B/32 is used here and we report the average performance of linear probe and unsupervised image retrieval.}
	\label{tbl:random}
	\centering
	\resizebox{0.9\linewidth}{!}{
		\begin{tabular}	{l  |  c  c  c  c|  c    c  c| c c}
			\toprule
			Tasks                 & \multicolumn{4}{c|}{Random Class Ratio ($r_1$)} & \multicolumn{3}{c|}{Random Feature Ratio ($r_2$)} & \multicolumn{2}{c}{Dropout Ratio ($r_3$)}                                                       \\

			                      & $ 0.05$                                         & $ 0.1$                                            & $ 0.3$                                    & $ 1.0$ & $ 1.0$ & $ 0.5$ & $ 0.25$ & $0.25$ & $0.5$ \\\midrule
			Linear Probe          & 85.1                                            & 85.7                                              & 84.9                                      & 77.9   & 85.7   & 85.5   & 84.2    & 85.4   & 85.1  \\
			Unsup. Retr.          & 62.3                                            & 62.8                                              & 62.1                                      & 55.9   & 62.8   & 62.7   & 62.0    & 62.5   & 62.3  \\ \midrule
			Unsup. Retr. $^{256}$ & -                                               & -                                                 & -                                         & -      & 61.4   & 61.8   & 61.0    & 60.7   & 60.1  \\
			\bottomrule
		\end{tabular}
	}
\end{table}

\noindent{\bf \textcolor{black}{Fine-tune for ImageNet-1K Classification.}}
In Tab.~\ref{tab:imagenet_results}, we compare our method with state-of-the-art supervised and weakly supervised pre-training \citep{dosovitskiy2021image,zhai2022scaling,singh2022revisiting} in transfer-learning experiments on ImageNet-1k. 
Our models consistently outperform OPEN-CLIP models in the Top-1 accuracy, showing the superiority of the proposed cluster discrimination. 
For ViT B/16, our pre-training achieves 85.9\%, surpassing both the supervised pre-training on IN-21K (84.0\%) and the weakly supervised pre-training on IG 3.6B (85.3\%). 
In addition, our ViT L/14 obtains 88.3\%, outperforming ViT L/16 pre-trained on JFT 300M (87.8\%) and ViT L/16 pre-trained on IG 3.6B (88.1\%). 
The overall results in ImageNet-1K classification task show that our models are very competitive as they can achieve better or comparable accuracy even though the training data used by the competitors are much larger (e.g., JFT 3B and IG 3.6B).

\noindent{\bf Fine-tune for Supervised Image Retrieval.}
In Tab.~\ref{tbl:sota}, we compare the proposed approach with the latest image retrieval methods \citep{Patel2022CVPR,Ermolov_2022_CVPR} trained with vision transformer. During fine-tuning of our models, the random negative class selection ratio $r_1$ is set to $1.0$ as the training data is clean.
Under different computing regimes, the proposed method consistently surpasses RA@K \citep{Patel2022CVPR} on the SOP, iNaturalist, and VehicleID datasets and outperforms Hyp-ViT \citep{Ermolov_2022_CVPR}  on the CUB and In-shop datasets.

\subsection{Ablation Study}

\noindent{\bf Encoder for Clustering.} In Tab.~\ref{tbl:abalationclusterand1strandom}, we compare the results of linear probe and unsupervised image retrieval under image-based clustering and text-based clustering by using the pre-trained CLIP and OPEN-CLIP models. As we can see, the text encoder is more powerful than the image encoder, and image and text signals are complementary as the joint clustering significantly outperforms each individual. \textcolor{black}{By referring to Tab.~\ref{tab:linear-probe-big-table} and Tab.~\ref{tab:zeroshotimageretrieval}, the OPEN-CLIP ViT B/32 model achieves $82.1\%$ and $55.2\%$ average results on the linear probe and unsupervised image retrieval tasks, while the proposed cluster discrimination method ($r_1=0.1$) significantly boosts the performance to $84.1\%$ and $61.1\%$ by using the OPEN-CLIP image and text models for clustering.} By using the CLIP image and text models for clustering, the performance can further increase to $85.7\%$ and $62.8\%$ on the linear probe and unsupervised image retrieval tasks. Therefore, we choose the CLIP models for clustering.

\noindent{\bf Cluster Number.}
In Tab.~\ref{tbl:abalationclusterand1strandom}, we also compare the performance under different cluster numbers, i.e., 100K, 1M, and 10M, by using the CLIP image and text models. As can be seen, the best results can be achieved when the cluster number is set as 1 million, with the average image number per class being around 400. \textcolor{black}{The cluster number needs to be balanced between the intra-class noises and inter-class noises.} Too small cluster numbers will incur heavy intra-class noise, which dramatically decreases the performance of the pre-trained classification model. Besides, too many clusters will increase the computation and storage burden on the FC layer. Most important, the over-decomposition will increase the inter-class noise ratio and undermine the discriminative power of the pre-trained model. 

\noindent{\bf Random Class Selection.} In Tab.~\ref{tbl:random}, we train ViT B/32 models under different inter-class sampling ratios. The basic margin-based softmax loss ($r_1=1.0$) only achieves $77.9\%$ on the linear probe task as it can hardly adapt to the heavy inter-class conflict in the automatically clustered dataset. When the sampling ratio is decreased from $1.0$ to $0.3$ and $0.1$, our method exhibits consistently better performance than the baseline, indicating random inter-class sampling is beneficial for the model's robustness. When $r_1$ is set to $0.05$, there is a slight performance drop because the inter-class interaction is insufficient during training. Therefore, we choose the random negative class selection ratio as $0.1$, obtaining $85.7\%$ and $62.8\%$ on the linear probe and unsupervised image retrieval tasks. 

\noindent{\bf Random Feature Selection.} 
In Tab.~\ref{tbl:random},  we compare the performance of the proposed Unicom under different random feature selection ratios ($r_2$) on the task of dimension-constrained unsupervised image retrieval.
\textcolor{black}{Here, we also include Dropout at different drop ratios ($r_3$) for comparison. From the results, we can have the following observations: (1) both random feature selection and Dropout can not improve linear probe and unsupervised image retrieval at a full dimension of $512$ as the LAION 400M dataset is large enough and regularization is not necessary for the final classification layer, (2) there is slight performance drop when the random feature selection ratio is decreasing, and (3) the proposed random feature selection ($r_2=0.5$) can improve $0.4\%$ for 256-D unsupervised image retrieval, while Dropout can not improve dimension-constrained unsupervised image retrieval. Even though Dropout enforces partial features for classification, the global randomization within the mini-batch makes the optimization involve all feature dimensions.
By contrast, the proposed random feature selection is fixed within the mini-batch thus it can benefit from optimization in a sub-feature space. }

\section{Conclusions}
This paper introduces Unicom, a simple yet effective framework for universal and compact feature embedding. Given the automatically clustered large-scale data, we employ one random negative class selection to improve the robustness under the heavy inter-class conflict and another random feature selection to improve the compactness of the feature representation. For both unsupervised and supervised image retrieval on different datasets, the proposed Unicom achieves state-of-the-art performance, confirming that cluster discrimination is beneficial to explore the semantic structure within the large-scale training data.

\bibliography{iclr2023_conference}
\bibliographystyle{iclr2023_conference}

\appendix
\section{Appendix}

\subsection{Model Architectures}
We follow the same architecture design as CLIP. Tab.~\ref{networkstructure} describes the details of architectures.

\subsection{Visualization of Pseudo Clusters and Data Distribution}
\begin{figure}
\centering
\subfigure[k = 100,000]{
\label{fig:100ka}
\includegraphics[width=0.32\textwidth]{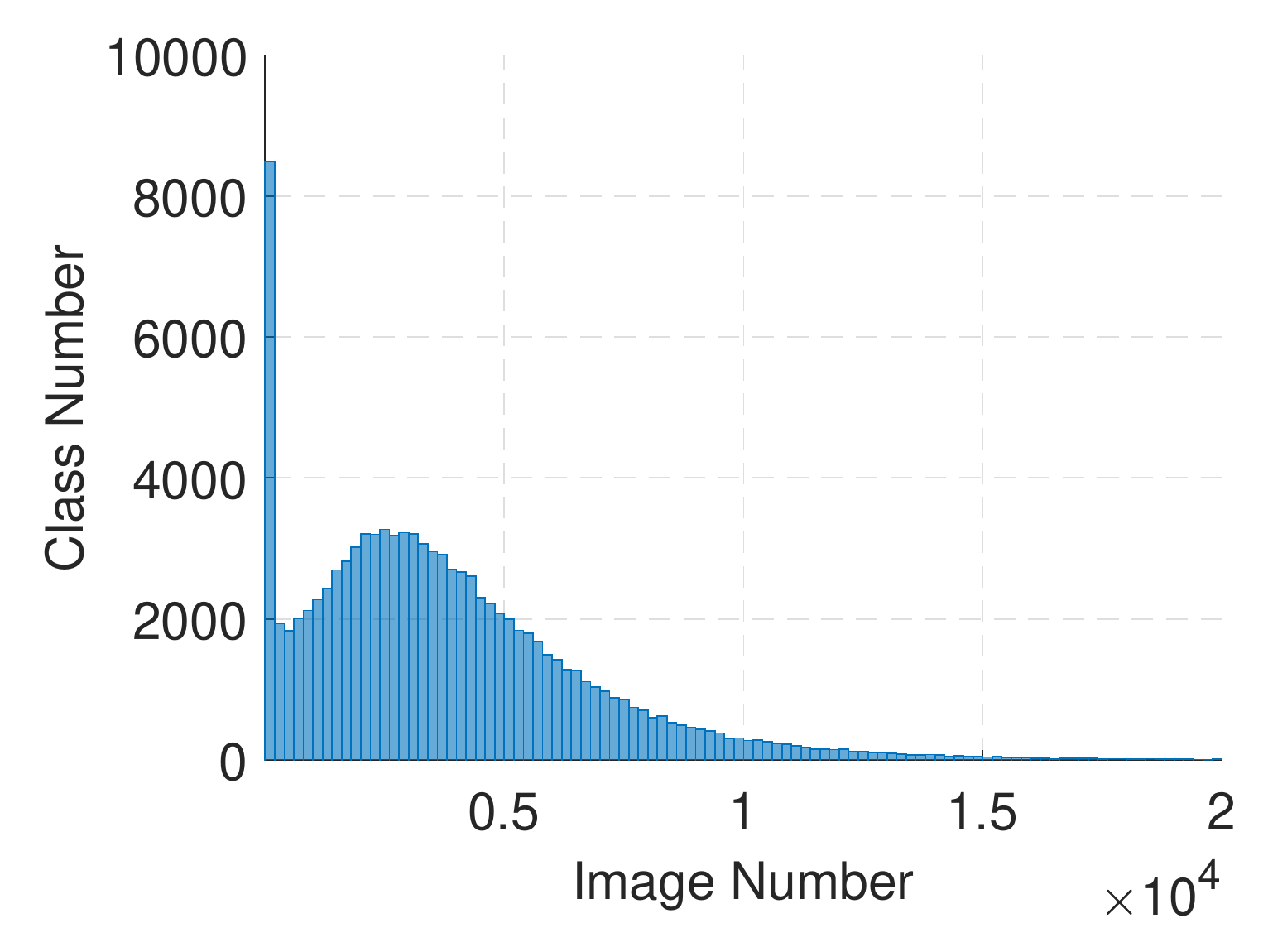}}
\subfigure[k = 1000,000]{
\label{fig:1Ma}
\includegraphics[width=0.32\textwidth]{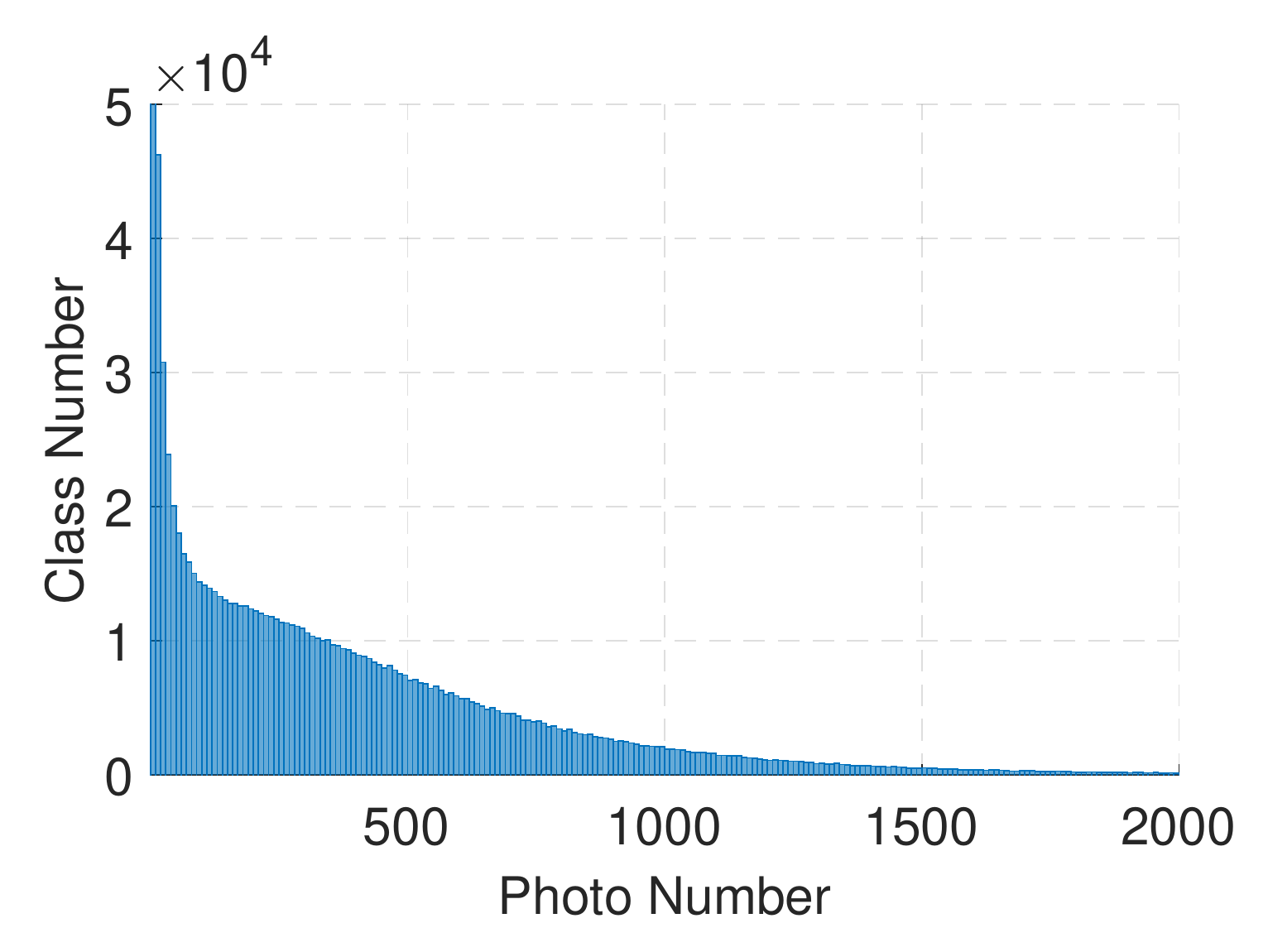}}
\subfigure[k = 10,000,000]{
\label{fig:10Ma}
\includegraphics[width=0.32\textwidth]{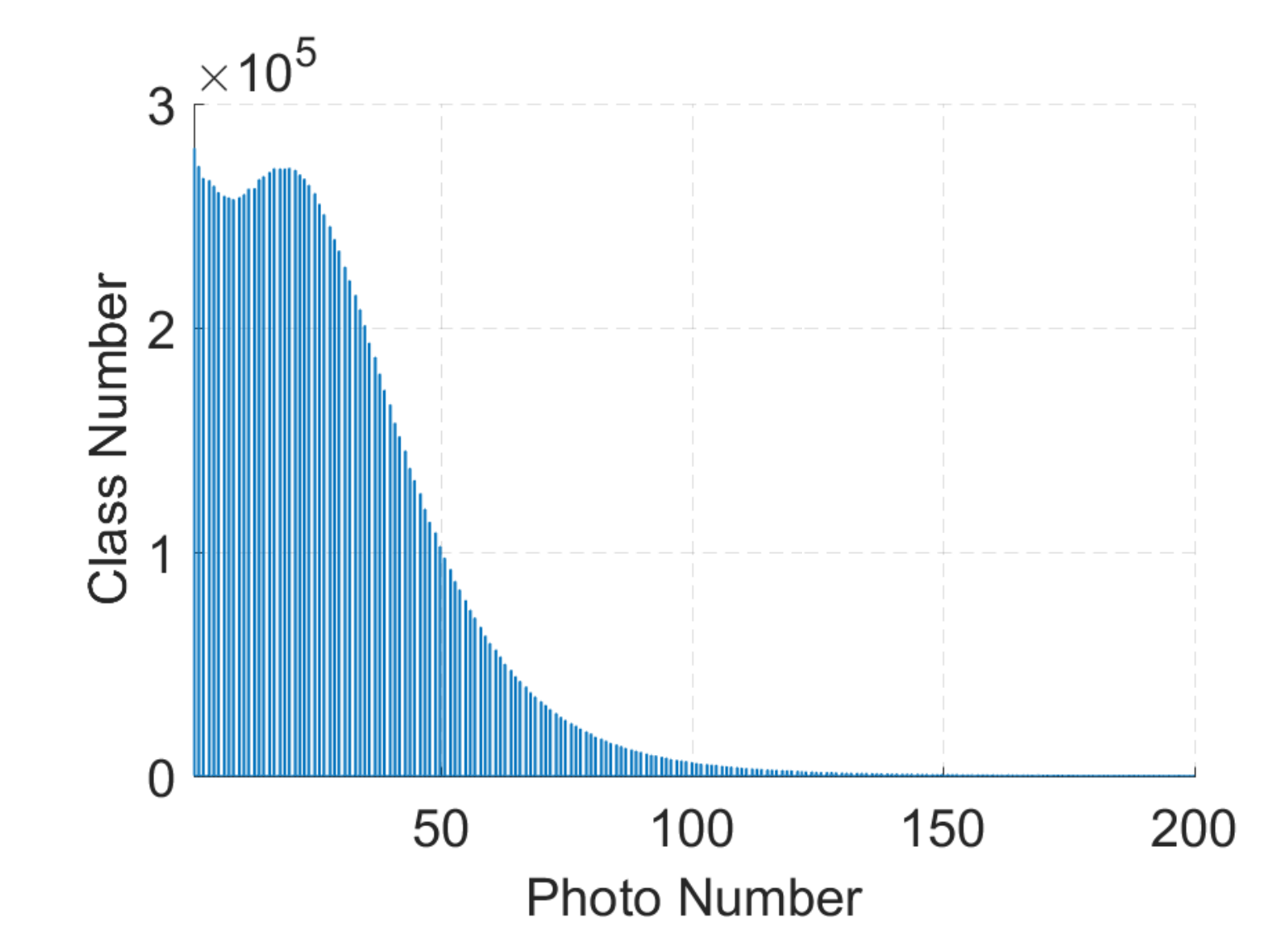}}
\caption{\textcolor{black}{Data distribution under different settings of cluster number $k$.}}
\vspace{-6mm}
\label{fig:kmeansa}
\end{figure}

\begin{figure}
\centering
\subfigure[Bag+Monkey]{
\label{fig1:bag1a}
\includegraphics[width=0.28\textwidth]{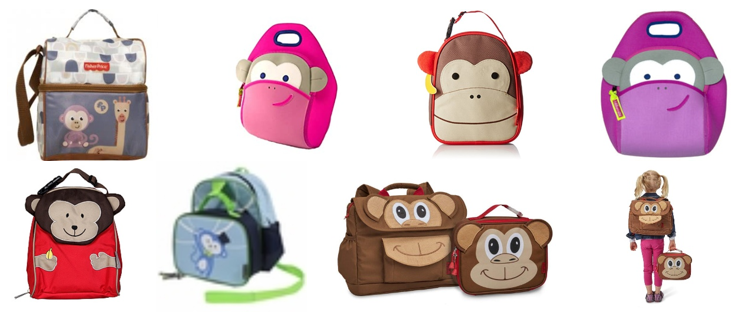}}
\hfill
\subfigure[Bag+Wing]{
\label{fig1:bag2a}
\includegraphics[width=0.28\textwidth]{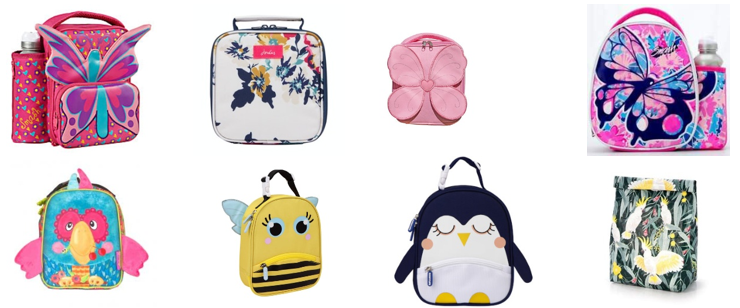}}
\hfill
\subfigure[Bag+Fantasy]{
\label{fig1:bag3a}
\includegraphics[width=0.28\textwidth]{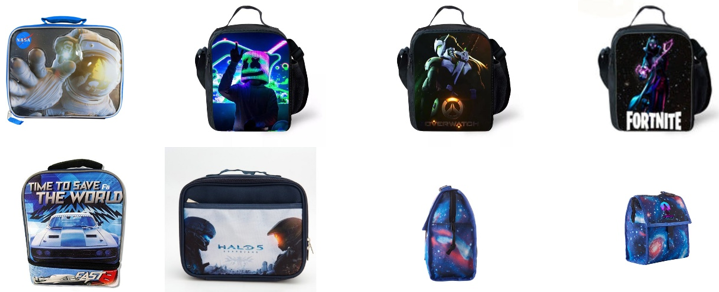}}

\subfigure[Top+Love]{
\label{fig1:top1a}
\includegraphics[width=0.28\textwidth]{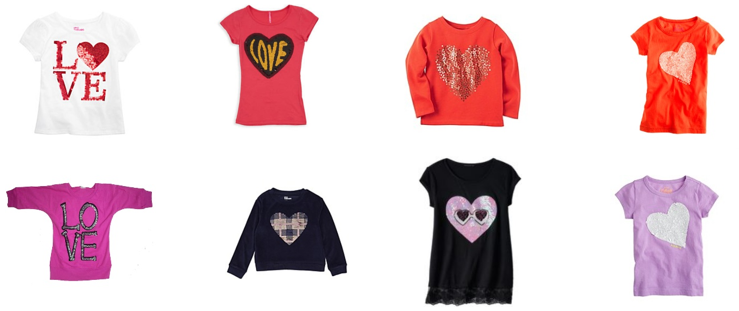}}
\hfill
\subfigure[Grey Top+Pink Sleeve]{
\label{fig1:top2a}
\includegraphics[width=0.28\textwidth]{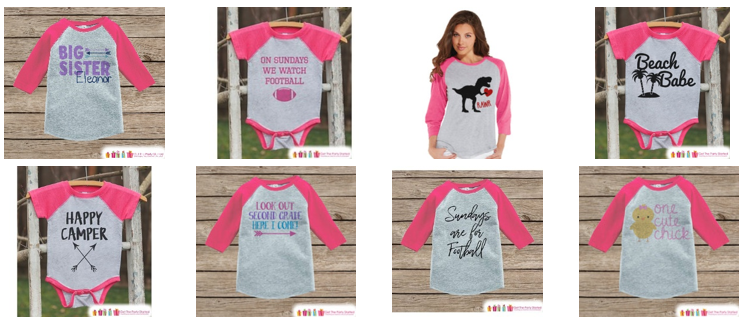}}
\hfill
\subfigure[Top+Cartoon]{
\label{fig1:top3a}
\includegraphics[width=0.28\textwidth]{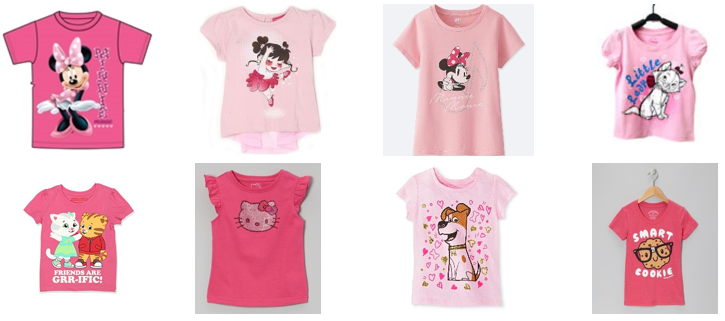}}
\vspace{-2mm}
\caption{Pseudo classes clustered by the image encoder and text encoder of the pre-trained CLIP model. The class name is given based on the manual observation of images and texts.}
\vspace{-6mm}
\label{fig1:cluteringappedixa}
\end{figure}

In Fig.~\ref{fig:kmeansa}, we show the data distribution under different settings of class number $k$.
In Fig.~\ref{fig1:cluteringappedixa}, we show some exemplar classes from the proposed automatic clustering.
As we can see, there are some fine-grained classes, such as the top with love icons and the top with cartoons.
Even though such clustering is reasonable and explainable, there is class confusion if we classify these samples from other views, such as color and targeting customer age.

\begin{table}
	\caption{The architecture parameters for ViT models.}
	\label{networkstructure}
	\begin{center}
		\begin{tabular}{l|c|c|c|c|ccc}
			\toprule
			Model        & Batch Size & FLOPs & Embedding & Input      & \multicolumn{3}{c}{Vision Transformer}                 \\
			             & (128 V100) & G     & dimension & resolution & layers                                 & width & heads \\
			\midrule
			ViT-B/32     & 256*128    & 4.3   & 512       & 224        & 12                                     & 768   & 12    \\
			ViT-B/16     & 256*128    & 17.6  & 768       & 224        & 12                                     & 768   & 12    \\
			ViT-L/14     & 64*128     & 80.9  & 768       & 224        & 24                                     & 1024  & 16    \\
			ViT-L/14-336 & 48*128     & 191.3 & 768       & 336        & 24                                     & 1024  & 16    \\
			\bottomrule
		\end{tabular}
	\end{center}
\end{table}

\begin{figure}
\centering
\includegraphics[width=0.5\textwidth]{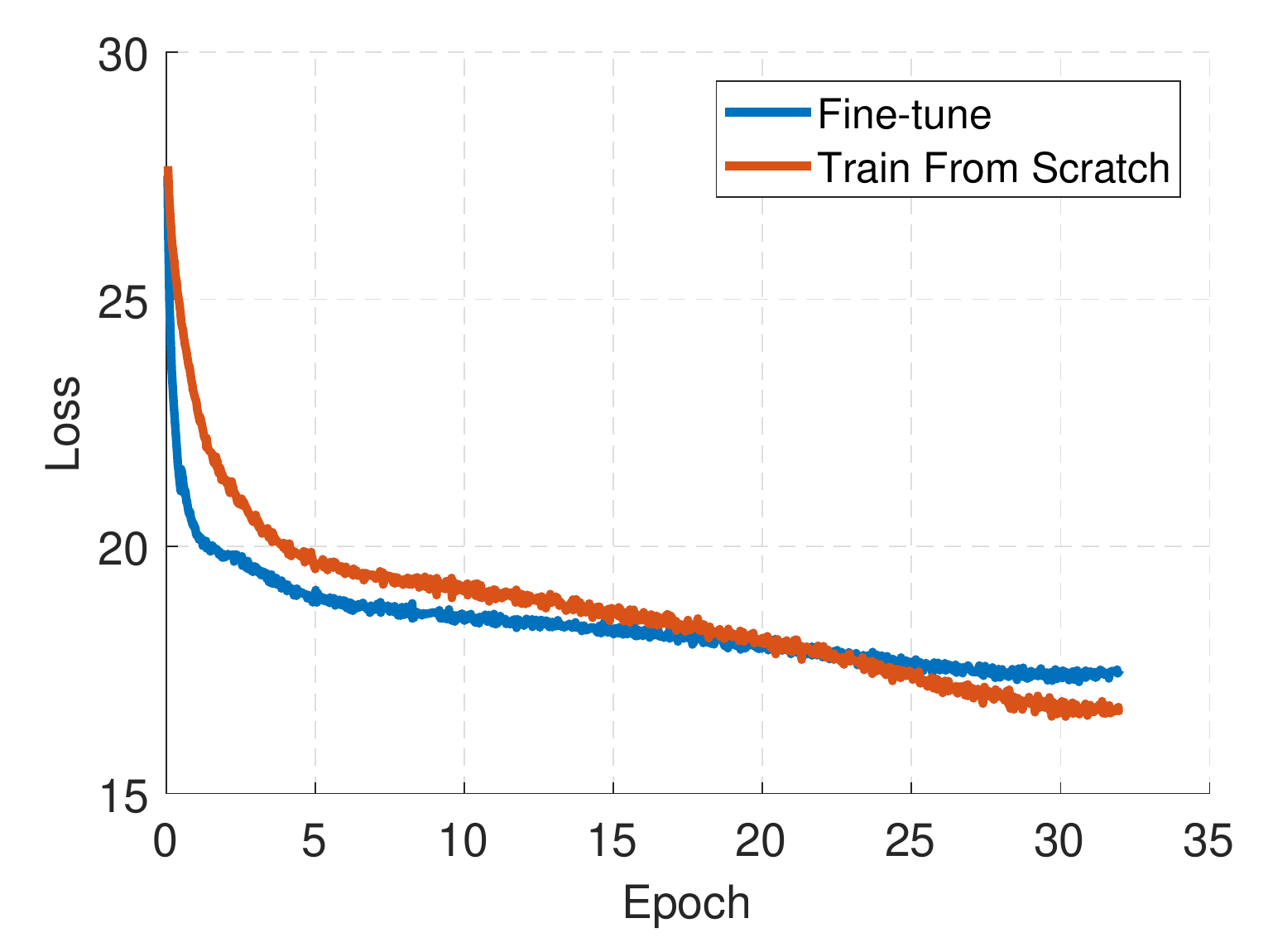}
\caption{\textcolor{black}{Training loss curves of fine-tuning from the CLIP model and training from scratch.}}
\label{fig:finetune}
\end{figure}

\begin{table}
	\caption{\textcolor{black}{Comparison between fine-tuning and training from scratch.} Top-1 accuracy(\%) of linear probe is reported on 13 image classification datasets. ViT B/32 is used here.}
	\label{tab:finetunevsscratch}
	\centering
	\resizebox{1.0\linewidth}{!}{
		\begin{tabular}{l|cccccccccccccc|c}
			\toprule
			                     &  & \rotatebox[origin=lb]{90}{\smash{CIFAR10}} & \rotatebox[origin=lb]{90}{\smash{CIFAR100}} & \rotatebox[origin=lb]{90}{\smash{Caltech101}} & \rotatebox[origin=lb]{90}{\smash{Cars}} & \rotatebox[origin=lb]{90}{\smash{Flowers}} & \rotatebox[origin=lb]{90}{\smash{Food101}} & \rotatebox[origin=lb]{90}{\smash{Birdsnap}} & \rotatebox[origin=lb]{90}{\smash{SUN397}} & \rotatebox[origin=lb]{90}{\smash{DTD}} & \rotatebox[origin=lb]{90}{\smash{Aircraft}} & \rotatebox[origin=lb]{90}{\smash{Pets}} & \rotatebox[origin=lb]{90}{\smash{EuroSAT}} & \rotatebox[origin=lb]{90}{\smash{ImageNet}} & \rotatebox[origin=lb]{90}{\smash{Average}} \\
			\midrule

			ViT B/32 (Scratch)   &  & 96.8                                       & 86.6                                        & 94.6                                          & 93.3                                    & 98.5                                       & 85.8                                       & 70.2                                        & 74.6                                      & 78.0                                   & 70.7                                        & 93.1                                    & 96.8                                       & 75.0                                        & 85.7                                       \\
			ViT B/32 (Fine-tune) &  & 95.8                                       & 83.3                                        & 94.1                                          & 92.4                                    & 98.5                                       & 87.3                                       & 67.3                                        & 75.3                                      & 79.8                                   & 66.5                                        & 92.8                                    & 96.1                                       & 75.1                                        & 85.0                                       \\
			\bottomrule
		\end{tabular}
	}
\end{table}

\begin{table}[t]
	\caption {Object detection and instance segmentation on COCO. We evaluate bounding-box AP (AP$^\text{bb}$) and mask AP (AP$^\text{mk}$) on val2017.}
	\label{tbl:coco}
	\centering
	\begin{tabular}	{l  |c|  c  c  c | c  c  c}
		\toprule
		Method                           & Pre-training Data & AP$^\text{bb}$ & AP$^\text{bb}_{50}$ & AP$^\text{bb}_{75}$ & AP$^\text{mk}$ & AP$^\text{mk}_{50}$ & AP$^\text{mk}_{75}$ \\
		\midrule
		ViT B/16 \citep{li2022exploring} & IN-1K             & 47.6           & -                   & -                   & 42.4           & -                   & -                   \\
		ViT B/16 \citep{li2022exploring} & IN-21K            & 47.8           & -                   & -                   & 42.6           & -                   & -                   \\
		ViT B/16 OPEN-CLIP               & LAION 400M        & 48.1           & 69.1                & 51.7                & 42.6           & 66.4                & 44.3                \\
		ViT B/16 Ours                    & LAION 400M        & 48.5           & 69.8                & 52.4                & 42.9           & 66.9                & 45.9                \\
		\bottomrule
	\end{tabular}
\end{table}
\subsection{\textcolor{black}{Training from Scratch vs. Fine-tuning from the CLIP Model}}
\textcolor{black}{
In this paper, the CLIP model is only used for the clustering step and our models are trained from scratch. In Fig.~\ref{fig:finetune}, we compare the training loss curves between fine-tuning and training from scratch. For fine-tuning, the backbone is initialized from the CLIP model (ViT-B/12), and the classifier (FC layer) is randomly initialized. The fine-tuning strategy can converge faster than training from scratch, but the final loss value is higher. In Tab.~\ref{tab:finetunevsscratch}, we also find that training from scratch outperforms fine-tuning from the CLIP model by $0.7\%$ on the task of linear probe.}

\subsection{Comparisons on COCO Detection and Segmentation}
Following the experiment setting in~\citep{li2022exploring}, we use Mask R-CNN~\citep{he2017mask} for bounding-box object detection and instance segmentation.
We fine-tune models on the COCO~\citep{lin2014microsoft} train2017 split and evaluate on the val2017 split.  
In Tab.~\ref{tbl:coco}, our method outperforms both OPEN-CLIP and supervised pre-training in all metrics, demonstrating the effectiveness of the proposed cluster discrimination.

\subsection{Linear Probe Datasets}
We use 13 image classification datasets to prove the effectiveness of our method. These
datasets include CIFAR10\citep{krizhevsky2009learning}, CIFAR100\citep{krizhevsky2009learning}, Caltech101\citep{fei2004learning}, Stanford Cars\citep{krause2013collecting}, Oxford Flowers\citep{nilsback2008automated}, Food-101\citep{bossard2014food}, Birdsnap\citep{berg2014birdsnap}, SUN397\citep{xiao2010sun}, Describable Textures\citep{cimpoi2014describing}, FGVC Aircraft\citep{maji2013fine}, Oxford-IIIT Pets\citep{parkhi2012cats}, EuroSAT\citep{helber2019eurosat}, ImageNet-1k\citep{ILSVRC15}. 
Details on each dataset and the corresponding evaluation metrics are provided in Tab.~\ref{linearprobedatasets}.

\subsection{Image Retrieval Datasets}
The training and evaluation of image retrieval experiments on seven widely used datasets, namely CUB-200-2011(CUB)~\citep{wbm+10}, Stanford Cars(Cars196)~\citep{ksd+13}, Stanford Online Products(SOP)~\citep{oh2016deep}, In-shop Clothes Retrieval(In-Shop)~\citep{liu2016deepfashion}, iNaturalist~\citep{van2018inaturalist}, VehicleID~\citep{liuhongye2016vehicles}, and Google Landmarks
dataset (GLDv2)~\citep{weyand2020google}. 
The number of examples and classes can be found in Tab.~\ref{tab:datasets}.

\begin{table}
	\caption{List of linear probe datasets with the data distribution and evaluation metrics.}
	\label{linearprobedatasets}
	\begin{center}
		\begin{tabular}{lrrrr}
			\toprule
			\multicolumn{1}{l}{\bf Dataset} & \multicolumn{1}{c}{\bf Classes} & \multicolumn{1}{c}{\bf Train size} & \multicolumn{1}{c}{\bf Test size} & \multicolumn{1}{c}{\bf Evaluation metric} \\
			\midrule
			CIFAR-10                        & 10                              & 50,000                             & 10,000                            & accuracy                                  \\
			CIFAR-100                       & 100                             & 50,000                             & 10,000                            & accuracy                                  \\
			Caltech-101                     & 102                             & 3,060                              & 6,085                             & mean-per-class                            \\
			Stanford Cars                   & 196                             & 8,144                              & 8,041                             & accuracy                                  \\
			Oxford Flowers                  & 102                             & 2,040                              & 6,149                             & mean per class                            \\
			Food-101                        & 102                             & 75,750                             & 25,250                            & accuracy                                  \\
			Birdsnap                        & 500                             & 42,283                             & 2,149                             & accuracy                                  \\
			SUN397                          & 397                             & 19,850                             & 19,850                            & accuracy                                  \\
			Describable Textures            & 47                              & 3,760                              & 1,880                             & accuracy                                  \\
			FGVC Aircraft                   & 100                             & 6,667                              & 3,333                             & mean per class                            \\
			Oxford-IIIT Pets                & 37                              & 3,680                              & 3,669                             & mean per class                            \\
			EuroSAT                         & 10                              & 10,000                             & 5,000                             & accuracy                                  \\
			ImageNet                        & 1000                            & 1,281,167                          & 50,000                            & accuracy                                  \\
			\bottomrule
		\end{tabular}
	\end{center}
\end{table}

\begin{table}
	\caption{Dataset composition for training and evaluation in the image retrieval task.}
	\label{tab:datasets}
	\begin{center}
		\begin{tabular}{lrr}
			\toprule
			\textbf{Dataset}                              & \textbf{Images} & \textbf{Classes} \\
			\midrule
			CUB Train~\citep{wbm+10}                      & 5,864           & 100              \\
			CUB Test ~\citep{wbm+10}                      & 5,924           & 100              \\
			Cars196 Train~\citep{ksd+13}                  & 8,054           & 98               \\
			Cars196 Test~\citep{ksd+13}                   & 8,131           & 98               \\
			SOP Train~\citep{oh2016deep}                  & 59,551          & 11,318           \\
			SOP Test~\citep{oh2016deep}                   & 60,502          & 11,316           \\
			In-Shop~\citep{liu2016deepfashion}            & 25,882          & 3,997            \\
			In-Shop~\citep{liu2016deepfashion}            & 26,830          & 3,985            \\
			iNaturalist Train~\citep{van2018inaturalist}  & 325,846         & 5,690            \\
			iNaturalist Test~\citep{van2018inaturalist}   & 136,093         & 2,452            \\
			VehicleID Train~\citep{liuhongye2016vehicles} & 110,178         & 13,134           \\
			VehicleID Test~\citep{liuhongye2016vehicles}  & 40,365          & 4,800            \\
			GLDv2 Train~\citep{weyand2020google}          & 1,580,470       & 81,314           \\
			GLDv2 Test ~\citep{weyand2020google}          & 762,884         & 1,129            \\
			\bottomrule
		\end{tabular}
	\end{center}
\end{table}

\end{document}